\documentclass{article}
\usepackage{kr}

\usepackage{times}
\usepackage{soul}
\usepackage{url}
\usepackage[hidelinks]{hyperref}
\usepackage[utf8]{inputenc}
\usepackage[small]{caption}
\usepackage{graphicx}
\usepackage{amsmath}
\usepackage{amsthm}
\usepackage{amsfonts}
\usepackage{amsopn}
\usepackage{amssymb}
\usepackage{booktabs}
\usepackage{algorithm}
\usepackage{algorithmic}
\usepackage{graphicx}  
\usepackage{float}  
\usepackage{subfigure}  

\title{Sample Efficient Imitation Learning via Reward Function Trained in Advance}

\author{%
Lihua Zhang$^1$\and
Quan Liu$^{1,2,3}$\\
\affiliations
$^1$School of Computer Science and Technology, Soochow University, Soochow\\
$^2$Provincial Key Laboratory for Computer Information Processing Technology, Soochow University, Soochow\\
$^3$Key Laboratory of Symbolic Computation and Knowledge Engineering of Ministry of Education, Jilin University, Jilin \\
\emails
lihua\_zhang92@163.com,
quanliu@suda.edu.cn
}

\begin{document}

\maketitle

\begin{abstract}
Imitation learning (IL) is a framework that learns to imitate expert behavior from demonstrations. Recently, IL shows promising results on high dimensional and control tasks. However, IL typically suffers from sample inefficiency in terms of environment interaction, which severely limits their application to simulated domains. In industrial applications, learner usually have a high interaction cost, the more interactions with environment, the more damage it causes to the environment and the learner itself. In this article, we make an effort to improve sample efficiency by introducing a novel scheme of inverse reinforcement learning. Our method, which we call \textit{Model Reward Function Based Imitation Learning} (MRFIL), uses an ensemble dynamic model as a reward function, what is trained with expert demonstrations. The key idea is to provide the agent with an incentive to match the demonstrations over a long horizon, by providing a positive reward upon encountering states in line with the expert demonstration distribution. In addition, we demonstrate the convergence guarantee for new objective function. Experimental results show that our algorithm reaches the competitive performance and significantly reducing the environment interactions compared to IL methods.
\end{abstract}

\section{Introduction}
Imitation Learning (IL), also known as Learning from Demonstration (LfD), is a framework which learning an optimal behavior policy from expert demonstrations. The expert demonstration generates by expert policy, and composed of state-action pair. Each state-action pair indicates the action to take at the state being visited. There are two ways to attaining the goal that leaning optimal policy from demonstrations, include Behavioral Cloning (BC) \cite{DBLP:conf/mi/BainS95} and Inverse Reinforcement Learning (IRL) \cite{DBLP:conf/icml/NgR00}.\\
BC methods learn an expert policy in a supervised fashion without environment interactions. Usually, behavioral cloning methods are the first option when have sufficient expert demonstrations \cite{DBLP:conf/iclr/SasakiYK19}. In the practical application, such as autonomous vehicles task, the environment usually has a vast majority of state-action pairs, however, the expert demonstrations only have a limited number of the states. Therefore, BC methods often suffer from compounding error \cite{DBLP:journals/jmlr/RossB10}, inaccuracies compound over time and can lead the learner to encounter unseen states in the expert demonstrations. Moreover, BC often can’t take the optimal action when encounters unseen states.\\
Inverse Reinforcement Learning (IRL) algorithm provide for an automated framework for decision making and control without reward function: by specifying a high-lever objective function, an IRL algorithm can, in principle, automatically learn a reward function and control policy that satisfies this objective. This has the potential to automate a range of applications, such as autonomous vehicles and robotic control. Given that the policy learns based on reward function, is the most succinct and robust way. Inverse Reinforcement Learning methods aim to recover reward or cost function by expert demonstrations, then IRL methods obtains optimal policy through a standard reinforcement learning algorithm. The target of the objective function is to train a reward function which guarantees the optimal policy trained by it as optimal as expert policy. IRL methods can take better action compared with BC algorithms, when encounters unseen states in the expert demonstrations. There for IRL can overcome the compounding error problem \cite{DBLP:conf/aaai/ZiebartMBD08}.\\
The original IRL framework utilizes a linear mapping between input features and out reward. These algorithms assume the unknown reward function that can be expressed as a linear combination of state-action pair’s features, this assumption severely restricts the complexity of reward structures that can be modeled accurately. and the reward function is difficult to apply to complex, high-dimensional tasks with large state-action space.\\
Finn and levine who proposed an algorithm that use expressive, nonlinear function approximators, such as neural networks, to represent the reward function \cite{DBLP:conf/icml/FinnLA16}. This approach learns nonlinear reward function from expert demonstrations, at the same time, the approach learns a policy to perform the task, and the policy optimization “guides” the reward toward good regions of the space.\\
To improve the representation ability of linear reward function, Levine proposed Gaussian Process Inverse Reinforcement Learning (GPIRL) \cite{DBLP:conf/nips/LevinePK11} that recovers the reward function with Gaussian Process (GP). GPIRL extends the GP model to account for the stochastic relationship between actions and underlying rewards. This allows GPIRL to balance the simplicity of the learned reward function against its consistency with the expert’s actions, without assuming the expert to be optimal. But GPIRL have a runtime dependent on the size of the dataset and are hard to processing large amounts of demonstration data.\\
There has some recent work on IRL algorithms that focus on recover reward function efficiently. Ho and Ermon who introduced a model-free imitation learning method called Generative adversarial Imitation Learning (GAIL) \cite{DBLP:conf/nips/HoE16}. This algorithm intimately connected to generative adversarial network \cite{DBLP:journals/corr/GoodfellowPMXWOCB14}, and does not interact with expert during training. The process of training generator and discriminator network is divided into two procedures: RL procedure and IRL procedure. IRL is a dual of an occupancy measure matching problem, in this procedure, discriminator manage to improve its ability to distinguish which action bring a policy’s occupancy measure closer to the expert’s. Unlike DAgger \cite{DBLP:journals/jmlr/RossGB11}, can simply ask the expert for such actions. \\
GAIL suffers from the problems of mode collapse and low sample efficiency in terms of environment interaction \cite{DBLP:conf/ijcai/LeHVNBP19}. The weakness of mode collapse is inheriting from GANs, and several works have built on GAIL to overcome this problem \cite{DBLP:conf/nips/LiSE17} \cite{DBLP:conf/ijcai/FeiWZZHZJL20}. Although, GAIL leverage the expert demonstrations efficiently, it needs a large number of environment interactions. Similar to GAIL, other model free inverse reinforcement learning algorithms are quite data-expensive to train, which often limits their application to simulated domains. These methods require solving the reinforcement learning procedure (finding an optimal policy given the current reward function) in the inner loop of an iterative reward optimization. This makes them need a majority of interaction with environment when apply to continuous control task with large state-action space, where the reinforcement learning procedure is difficult to train. Particularly real-world systems with unknown dynamics.\\
We are interested in imitation learning and want to address this issue, because we desire an algorithm that can be applied to real-word problems for which it is hard to design the reward, and the cost of interaction with environment is expensive. Furthermore, in most real-word problems, even if the expert safely demonstrated, the learner may have policies that damage the environments and the learner itself during training. In this paper we focus on model-free imitation learning for continuous control. \\
To address this problem, we propose a new approach to improve sample efficiency suffered by previous methods, which we call Model Reward Function Based Imitation Learning (MRFIL). The main contribution of our work is we dividing IRL algorithm into two steps: IRL procedure and RL procedure. In IRL procedure, MRFIL pre-trains a fixed reward function by expert demonstrations. In RL procedure, based on the pre-trained reward function, MRFIL learns an optimal policy with standard RL algorithms. We also propose a new objective function, and analyze the convergence theoretically. Unlike prior IRL methods, by pre-training a fixed reward function and learning an optimal policy in a single RL procedure, we considerably shrink the amount of interactions that interact with environment. Our evaluation demonstrates the performance of our method on a set of simulated benchmark tasks, showing that it achieves state-of-the-art performance as compared with a number of imitation learning task what difficult to interact to environment. 

\section{Related Work}

Currently, the problem of learning an optimal policy in sample efficient way is still not well understood. Sample efficiency imitation learning data to at least work of Guided Cost Learning (GCL), which updates the cost function in the inner loop of policy search. specifically, GCL directly optimizing a trajectory distribution with respect to the current cost function using sample-efficient reinforcement learning algorithm. Thus, the cost function can lead the learner toward regions where the samples are more useful, and inherits its sample efficiency. However, GCL require the model is well-approximated by iteratively fitted time-varying linear dynamics, and have worse imitation capability compared with GAIL \cite{DBLP:journals/corr/abs-1710-11248}.\\ 
GAIL is based on prior IRL works, and has achieved state-of-the-art performance on a variety of continuous control tasks, GAIL overcome the compounding error of BC methods, and is quite sample efficient than BC in terms of expert demonstration data. However, it suffers from the sample efficient in terms of environment interaction, which severe restricts its application in practical problems.\\ 
Hester and Osband proposed an approach use a prioritized replay mechanism to automate learn optimal policy\cite{DBLP:conf/aaai/HesterVPLSPHQSO18}. However, this algorithm needs expert demonstration and hand-crafted reward, we address the problem only need expert demonstration.\\
At present, the adversarial IL (AIL) framework has become a popular choice for IL \cite{DBLP:conf/icml/BaramACM17};  \cite{DBLP:conf/nips/HausmanCSSL17}; \cite{DBLP:conf/nips/LiSE17}. But these methods require a majority of state-action pairs obtained through the interaction between the learner and environment. To address this problem, Blondé and Kalousis leveraging an off-policy architecture to reduce interaction number with environment \cite{DBLP:conf/aistats/BlondeK19}. However, this algorithm still needs reward function to learn through interact with environment.\\ 
Fumihiro and Tetsuya adopt off-policy actor-critic (Off-PAC) algoritm \cite{DBLP:conf/iclr/SasakiYK19}. Compare this algorithm estimating the state-action value using off-policy samples without learning reward function. In contrast, our method trains reward function with offline.

\section{Background}

\subsection{Markov Decision Process}
We focus on model-free imitation learning for continuous control task in this work, and model this task as a Markov Decision Process (MDP), can be formalized by the tuple:$(S,A,p,r,\gamma ,\rho_{0})$. Where $ S $ is the state space, $ A $ is the action space, $ p $ is the unknown transition function, 
$ p:S\times A\times S \rightarrow [0,1] $, specifying the probability density $ p(s',r|s,a) $ to the next state $ s' $ acquire reward $ r $ from the current state $ s $ by taking action $ a $. $ r:S \times A \times S \rightarrow R $ is the reward function, $ \rho_{0}:S\rightarrow[0,1] $ is the distribution of initial states. A policy is a function $ \pi:S \times A \rightarrow [0,1] $, which outputs a distribution over the action space for a given state $ s $.
\subsection{Policy Gradient Methods}
In continuing tasks, where environment interactions are unbounded in sequence length, the returns $R_{t}$ at time step t for a trajectory are defined as $R_{t}=\sum_{k=t}^{\infty}\gamma^{k-t}r(s_{k},a_{k},s_{k+1})$. The policy gradient theorem assumes that every trajectory starts in some particular (non-random) state $s_{0}$, and $s_{0}$ leads learner to initial states according to distribution $\rho_{0}$. Then, the goal is to learn a policy that maximizes expected returns in state $s_{0}$.
		\begin{equation} \label{0}
				J(\boldsymbol{\theta}) \triangleq v_{\pi_\theta}(s_0)=\sum_{t=0}^{\infty}\gamma^{t}r(s_t,a_t,s_{t+1})	
		\end{equation}
where $v_{\pi_{\theta}}{(s_0 )}$ denotes the value of state $s_0$  under policy $\pi$.
Thus, the performance depends on both the action selections and the distribution of states in which those selection are made, and that both of these are affected by the parameter of policy network \cite{sutton2018reinforcement}.

\subsection{Inverse Reinforcement Learning Methods}
A common assumption in IRL is that the demonstrator utilizes a Markov Decision Process for decision making. Normally, IRL algorithms assumes the expert policy is optimal. The goal of IRL is to recover the unknown reward function from expert demonstrations \cite{DBLP:journals/ftrob/OsaPNBA018}. Recover the reward function can be beneficial when the reward function is the most parsimonious way to describe the desired behavior. \\
In IRL methods, the reward function is designed for reward function which makes the current policy as optimal as expert policy. The current policy is update using standard reinforcement learning algorithms based on the current estimator of the reward function. By repeating this procedure, the optimal policy and reward function can be obtained.

\section{Proposed Method}
Currently, Imitation learning methods obtains a policy as optimal as expert policy thorough at least millions interaction with environment. This severely limited imitation learning methods be widely utilized in the industrial field, due to the high interaction cost. Therefore, we introduce a noval imitation learning method that sucessfully addresses the impeding sample efficiency, in the number of interaction with environment.
\subsection{Algorithm Formulation}
It has various reasons that cause a low sample inefficiency. First, IL learner lacks of prior knowledge about environment, thus, all the information and knowledge that assist learner to take an action when visit a state can only be acquired by interact with environment. Secondly, based on current reward function, IRL methods alternate execute RL procedure and IRL procedure , this hugely increased the interaction number. Last, when human being learn a skill by imitating expert, the skill of human improves faster by learning a few expert demonstrations and samples. However, comparing with human, IL agent requires a majority of expert demonstrations and samples. One of the reason is that human being learn a new task with Prior-Knowledge.\\
Based on the above analysis, in order to reduce the number of interact with environment, MRFIL performs three important modifications.
\begin{enumerate}
	\item	Training an ensemble of dynamic model and a single dynamic model with expert demonstration data, then we use the variance predicted by an ensemble of dynamic model which composed of five dynamic models as a reward. The motivation is that the prediction of ensemble dynamic model in expert demonstration sate-action space tends to certainty, conversely, and the prediction of ensemble dynamic model far from expert demonstration sate-action space tends to uncertainty. Consequently, according to this characteristic, the reward function could identify whether the visited state belong to expert demonstrations, and the reward function is acquired before algorithm interacting with environment, and the policy can be optimized through reinforcement learning method. Besides, a single dynamic mode is used to pre-process neural network.
	\item	In order to reduce the number that agent interactions with environment, we use a single dynamic model and expert demonstration data to train algorithm in an offline way. We pre-train the algorithm with multi branch and short rollout, and the rollouts is divided to two ways: exploration rollout and exploitation rollout. This pretreatment provides a regularizing effect before policy learning, and is helpful to reduce training time.
	\item	Our theoretical result show that, MRFIL will cannot guarantee the convergence, if still adopt traditional objective function used in RL methods. To solve this problem, we add a supervised loss term to the objective function, and the supervised loss term is as same as BC loss. Furthermore, we demonstrate that the modified objective function guarantees the constringency in our approach. 
\end{enumerate}

\subsection{Ensemble Reward Function}
Generally, inverse reinforcement learning algorithms recover reward function with a single neural network, and learn it iteratively. Such as the popular choice for now: adversarial imitation learning. \\
We simplify this process to two single procedure: IRL procedure and RL procedure. In IRL procedure, MRFIL pre-trains a reward function. The reward function is defined as an ensemble of dynamic model, and the enssemble dynamic model is $ \boldsymbol{M}={m_{1},m_{2},…,m_{n}} $. We use expert demonstrations to train the ensemble dynamic model via standard supervised learning method, each single dynamic model is only differ by the initial weights. In this way, the reward function can be acquired before the RL procedure start, and the optimal policy can be learned by standard reinforcement learning method.\\
The input of dynamic model is state-action pair $ (s,a) $, the output is next state $ s' $. After the training procedure finished, if the current state-action pair belong to expert demonstrations, the variance of ensemble dynamics will be small. Conversely, if the current state-action pair far from expert state-action pair, the variance of ensemble dynamics will be large. Based on this principle, we define a hyper-parameter $ T_h $, to measure whether the input state-action pair belong to expert demonstrations. Concretely, the hyper-parameter is defined on the variance the output of ensemble dynamic models due to Pinsker’s inequality. The ensmble reward function is as follows:
		\begin{align} \label{1}
			var( \boldsymbol{M}(s,a))=var(m_{1}(s,a),\cdots,m_{n}(s,a))
		\end{align}

		\begin{align} \label{2}
			 r(s,a)= \left\{
			\begin{array}{rl}
				1 & \text{if} \  var( \boldsymbol{M} (s,a))  > T_{h}, \\
				0 & \text{else}.
			\end{array} \right. 
		\end{align}%
This modified reward function encourages the learner to explore in the state-action space close to expert demonstrations, and the learner is capable to go back when it far from the expert sate-action space. 
\subsection{Pre-training with Dynamic Model}
We adopt actor-critic frame to learn optimal policy in RL procedure. In the most of case, the algorithm converges finally. However, in the initial progress, the actor and critic network have a large variance which is derived from policy gradient methods that caused algorithm unstable and slower. Naturally, this problem leads to demand more samples to adjust learner policy. To address this problem, and improve sample efficiency, we pretrain the actor-critic network in an offline way. Same as ensemble dynamic model, we train a single dynamic model via expert demonstrations, which be used to predict next state, and the reward still provide by the ensemble dynamics.\\ 
The reason that caused a gap between sampling in true dynamic and dynamic model include two aspects: prediction error due to training data, and distribution shift due to the policy encountering states outside expert policy state-action space \cite{DBLP:conf/nips/JannerFZL19}.\\
In current settings, the dynamic model is a local optimal model, only accurate when encountering the state-action space belong to expert demonstrations. Even the local optimal cannot be guaranteed when expert demonstration is insufficient.\\ 
Based on the above analysis, we propose an offline pre-train algorithm via the expert demonstrations and dynamic model. In order to reduce the compound error with the increase of rollout length, we train the algorithm with multi branch and short rollout. The algorithm starts several rollouts under the state distribution of expert demonstrations, and the rollouts is divided to two types: exploration rollout and exploitation rollout.\\
In the exploration rollout procedure, learner acts with stochastic noise, and the distance that agent runs is closes. The noisy action leads learner to “sub-optimal” region mostly, which is unfamiliar to dynamic model. In this region the dynamic model has high prediction error, but the ensemble dynamic models provide zero reward in all regions that outside expert demonstrations. Therefore, the prediction error has no impact on the accuracy of learner. Furthermore, this provides a regularization effect during policy learning by penalizing policy that visit “sub-optimal” region \cite{DBLP:conf/nips/KidambiRNJ20}.\\
In the exploitation rollout procedure, learner act similar to standard model-based methods, but the distance that agent runs is far. This procedure in order to imitates expert policy in “optimal” region, and the dynamic model is familiar to this region thus has high prediction accuracy.\\
We propose a corollary to measure the performance of pre-trained learner, and the theoretical proof is given. This corollart could measure the gap between expert return in real environment and learner return in dynamic model. The expert return in real environment can be sampled in expert demonstrations, and the theoretically analysis presented in section 5.1. The more details about how to pre-train the IL agent are presented in algorithm 1.
	\begin{algorithm}[tb]
		\caption{Algorithm 1 MBSR: Multi Branch and Short Rollout Offline Pre-training}
		\label{alg:MBSR}
		\begin{algorithmic}[1] 
			\STATE \textbf{Require}: Expert demonstrations $D_{E}$
			\STATE Learn approximate dynamics model $ m_{0} $ and $\boldsymbol{M}={m_{1},m_{2},…,m_{n}}$, $ S \times A \rightarrow S $ using $ D_{E} $.
			\FOR {each branch $ b=1,2,… $}
			\FOR {each step of exploration branch}
			\STATE choose noisy a from $ s $ 
			\STATE $ m_{0} $  provide next state $ s' $, $ \textbf{M} $ provide reward $ r $
			\STATE update actor and critic network
			\ENDFOR	
			\FOR {each step of exploitation branch}
			\STATE choose noisy $ a $ from $ s $
			\STATE $ m_{0} $  provide next state $ s' $, $ \textbf{M} $ provide reward $ r $
			\STATE update actor and critic network
			\ENDFOR
			\ENDFOR
		\end{algorithmic}
	\end{algorithm}
\subsection{Imitation Learning with Ensemble Reward Function}	
In the RL procedure, we adopt Soft Actor-Critic (SAC) algorithm. If use the SAC objective function directly, the convergency of MRFIL will not guaranteed, and the theoretical analyses is given in section 5.2. Therefore, we modify the objective function by adding a supervise term about actor loss, this modification is as follows:
		\begin{equation} \label{3}
			\begin{split}
			\boldsymbol{\theta}_{t + 1} =\ & \boldsymbol{\theta}_{t} + \alpha \nabla_{\theta_t} \log  \pi	(a_t | s_t, \boldsymbol{\theta}_t) (r_{t+1} \\
			& + \gamma Q (s_{t+1},a_{t+1},\boldsymbol{\omega}) - Q(s_t,a_t,\boldsymbol{\omega})) \\
			& + \underbrace{\tau \nabla_{\theta_t}||\pi_E(a_t|s_t) - \pi (a_t|s_t,\boldsymbol{\theta}_t)||_2}_{\text{supervise term}}
			\end{split}		
		\end{equation}
where $ \boldsymbol{\theta}_t $ denotes the parameter of actor network, $ r_{t+1} $ denotes the reward acquired by take action $ a_{t} $ in state $ s_{t} $, which is provided by ensemble dynamics $ \boldsymbol{M} $. $ \alpha , \gamma , \tau $ denotes hyper-parameters.\\
The more details are presented in algorithm 2. Adding this supervise loss term enables the imitation policy in line 6 of Algorithm 2 reach to expert state-action space quickly, and learning around this optima region. Moreover, this modification guarantees the imitation policy converge to expert policy. We theoretically demonstrate this modification in section 5.2, and the convergence is proved by mathematical illation in \textbf{Theorem B3}.

\begin{algorithm}[tb]
	\caption{Algorithm 2 MRFIL: Model Reward Function Based Imitation Learning for Sample Efficient }
	\label{alg:MRFIL}
	\begin{algorithmic}[1] 
		\STATE \textbf{Input}: Expert demonstration data $ D_E=\left\{{(S_E,A_E,S^{'}_{E} )_{i} } \right \}_{i=1}^{n} $
		\STATE 	Initialize network weight $\boldsymbol{M}=\left\{m_1,m_2,…,m_n \right\} $ and \ $ \pi $,\ $\pi_E$, \ $Q$ 
		\STATE Initialize 	an empty replay pool $ D_{re} \leftarrow \emptyset $
		\STATE Use BC method to pretrain $ \pi_E,\boldsymbol{M}={m_1,m_2,…,m_n } $ with data $ D_E $
		\STATE Use Algorithm 2 to pre-train the actor-critic network			
		\WHILE{$ \pi $ and $ Q $ not converged} 
		\STATE Sample action $ a_t $ from the policy $ \pi $, and get transition $ s_{t+1} $from the environment 
		\STATE Get reward $ r_{t+1} $ from ensemble dynamic model
		$ D_{re} \leftarrow D_{re} \cup \left\{s_t,a_t,r_{t+1},s_{t+1}\right\} $
		Sample from replay pool to minimize policy gradient and value gradient 
		
		\begin{equation*} 
			\begin{split}
				\boldsymbol{\theta}_{t + 1} =\ & \boldsymbol{\theta}_{t} + \alpha \nabla_{\theta_t} \log  \pi	(a_t | s_t, \boldsymbol{\theta}_t) (r_{t+1} \\
				& + \gamma Q (s_{t+1},a_{t+1},\boldsymbol{\omega}) - Q(s_t,a_t,\boldsymbol{\omega}))    \\
				& + \tau \nabla_{\theta_t}||\pi_E(a_t|s_t) - \pi (a_t|s_t,\boldsymbol{\theta}_t)||_2   
			\end{split}								
		\end{equation*}
		
		\begin{equation*}
			\begin{split}						
				\boldsymbol{\omega}_{t+1}=& \boldsymbol{\omega}_{t} + (r_{t+1} + 
				\gamma \max Q(s_{t+1},a_{t+1},\omega_t)\\
				& - Q(s_{t},a_{t},\omega_t)\nabla_{\omega_t}Q(s_{t},a_{t},\omega_t) 
			\end{split}		
		\end{equation*}
		\ENDWHILE
		\STATE \textbf{Output} $\boldsymbol{\theta}$
	\end{algorithmic}
\end{algorithm}

\section{Theoretical Analysis}
This section provides the formal theoretical analysis of Model Reward Function Based Imitation Learning. 
\subsection{The Return Gap of Offline Pre-training}
Under such a scheme, we propose a corollary to measure the performance of pre-trained IL learner, the maximal gap between IL learner and expert can be obtined by Corollary A 1. The corollary is as follows: \\
\textbf{Corollary} A 1. (\textit{multi branch and short rollout}). \textit{Suppose the real dynamic transformation is} $ p_r (s'|s,a) $ , \textit{the dynamic model transformation is} $ p_m (s'|s,a) $.
		\begin{equation} \label{4}
				|\eta_{e} - \eta_{m}| \leq 2 \left(  \frac{\gamma(\epsilon_{m}+\epsilon_{\pi})}{(1-\gamma)^2}
				+ \frac{\epsilon_{\pi}}{1-\gamma} \right)				  
		\end{equation}
\textit{Proof}. See Appendix A, Corollary A.1.\\
$\eta_e$ denotes the returns of the expert policy in real dynamic MDP, $ \eta_m $ denotes the returns of the current policy in dynamic model MDP, $ \gamma $ denotes the hyperparameter, $ \pi_e $ denotes expert policy, $ \pi_m $  denotes the policy learned by pre-training. where $ \max_t \mathbb {E}_{s\sim p^t_m}(s)[D_{KL}(p_m(s'|s,a)||p_r(s'|s,a))] \leq \epsilon $, and $ \max_t D_{TV}(\pi_e(a|s)||\pi_m(a|s) \leq \epsilon_{\pi} $. The episode return is an indicator of learner performance, thus we the return gap $ \eta_{e}-\eta_{m} $  to measure the pre-training performance.\\

\subsection{Convergence of The Algorithm}
The objective function of inverse reinforcement learning algorithms is as follows:	
		\begin{equation} \label{5}
				\begin{split}
				\mathop{\mathrm{maximize}}_{r\in R} \Bigl( \min_{\pi} - & H(\pi) - \mathbb{E}_{\pi}[r(s,a)] \\
				+ & \mathbb{E_{\pi_{E}}}[r(s,a)] \Bigr) 	  
				\end{split}		
		\end{equation}	
where $H(\pi) \triangleq \mathbb{E}_{\pi}[-\log \pi(a|s)]   $ is $ \gamma $ discounted causal entropy of policy $ \pi$, $ \pi_E $ denotes expert policy.\\
\textbf{Theorem 1}. \textit{The dual of inverse reinforcement problem is as follows}: 
		\begin{equation} \label{6}
				\begin{split}
					\mathop{\mathrm{minimize}}_{\rho \in D} & \quad -H(\rho) \\
					\mathrm{s.t.} & \quad \rho(s,a)=\rho_{E}(s,a) \quad \forall s \in S,\ a \in A
				\end{split}		
		\end{equation}	
\textit{Proof}. See Appendix B, Theorem B.1.\\
Where we denote occupancy measure $  \rho_\pi:S\times A \rightarrow \mathbb{R} $ as $ \rho_{\pi}(s,a) = \pi(a|s)\sum_{t=0}^{\infty} \gamma^{t} P(s_{t} = s|\pi) $.\\
Theorem 1 implies that the goal of IRL is to find an optimal policy that the occupancy measure equals to expert policy. Following Theorem 1, that the learner taking action as same as expert in every state visitation can be guaranteed. \\ 
The objective function of policy gradient methods with entropy term, but without supervise term is as follows:
		\begin{equation} \label{7}
			\begin{split}
				\min_{\pi} \quad -H(\pi) - \mathbb{E}_{\pi}[r(s,a)]
			\end{split}		
		\end{equation}
\textbf{Theorem 2}. \textit{The dual problem of policy gradient methods objective function (Eq (7)) is as follows}:
		\begin{equation} \label{8}
			\begin{split}
				\mathop{\mathrm{minimize}}_{\rho \in D} & \quad -H(\rho) \\
				\mathrm{s.t.} & \quad \rho(s,a) \geq 0 \quad \forall s \in S,\ a \in A
			\end{split}		
		\end{equation}
\textit{Proof}. See Appendix B, Theorem B.2.\\
Theorem 2 implies that use SAC to learn optimal policy requiring the occupancy measure $ \rho(s,a) $ greater than zero, which is already satisfied. Thus, the reward function that learnes from expert demonstrations must be very precise, but this mathematical condition is hard to satisfy without interaction with environment. To address this problem, we modify the objective function Eq(\ref{7}), the modified objective function is as follows:
		\begin{equation} \label{9}
			\begin{split}
				\min_{\pi} \quad -& H(\pi) - \mathbb{E}_{\pi}[r(s,a)]\\
				 + & \underbrace{\mathbb{E}_{D_{E}}[\pi(s,a)-\pi_{E}(s,a)]}_{\text{supervise term}}
			\end{split}		
		\end{equation}

\begin{figure*}[ht]
	\centering  
	\subfigure{\label{sub.1} 
		\includegraphics[width=0.3\linewidth]{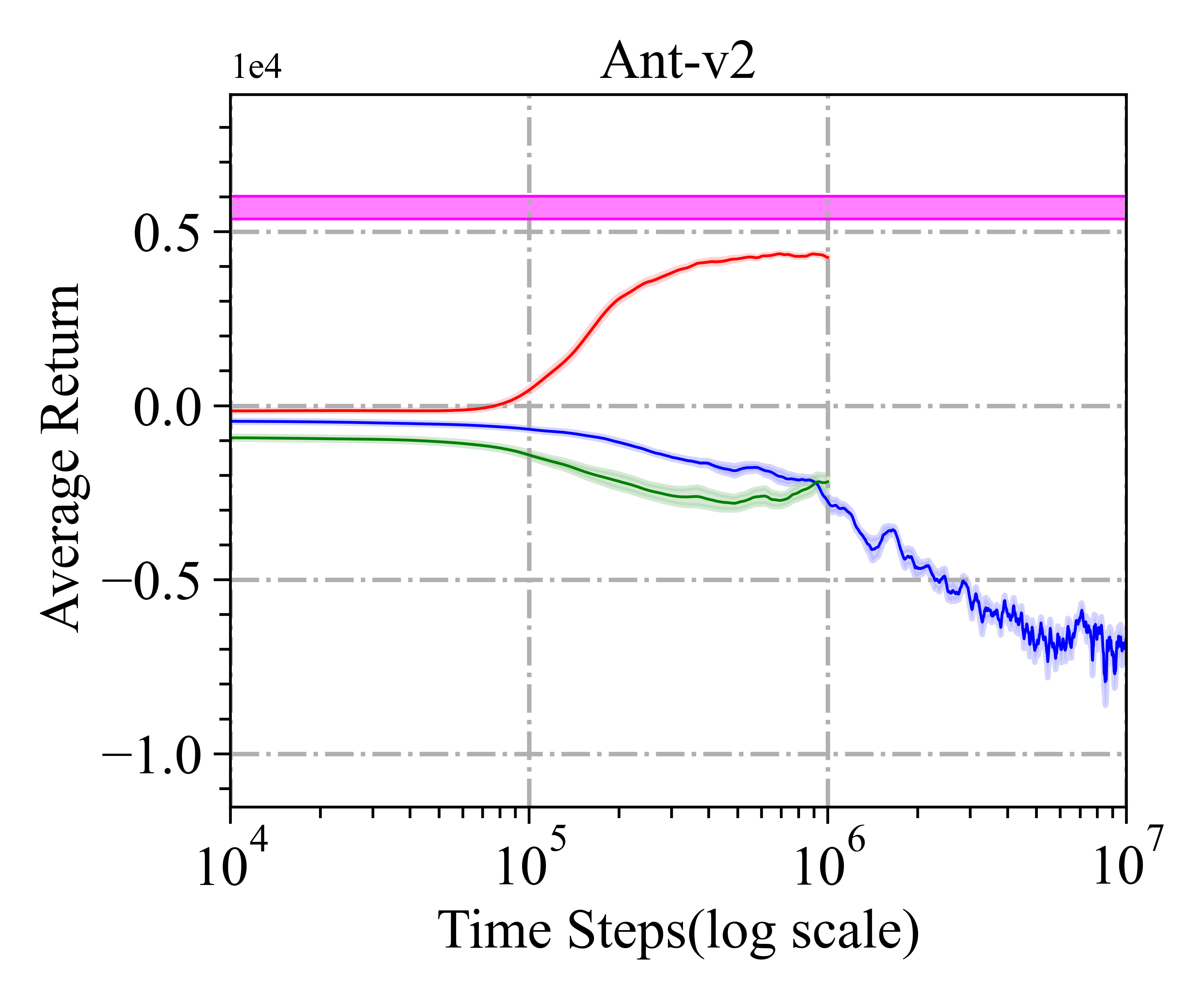}}
	\subfigure{\label{sub.2}
		\includegraphics[width=0.3\linewidth]{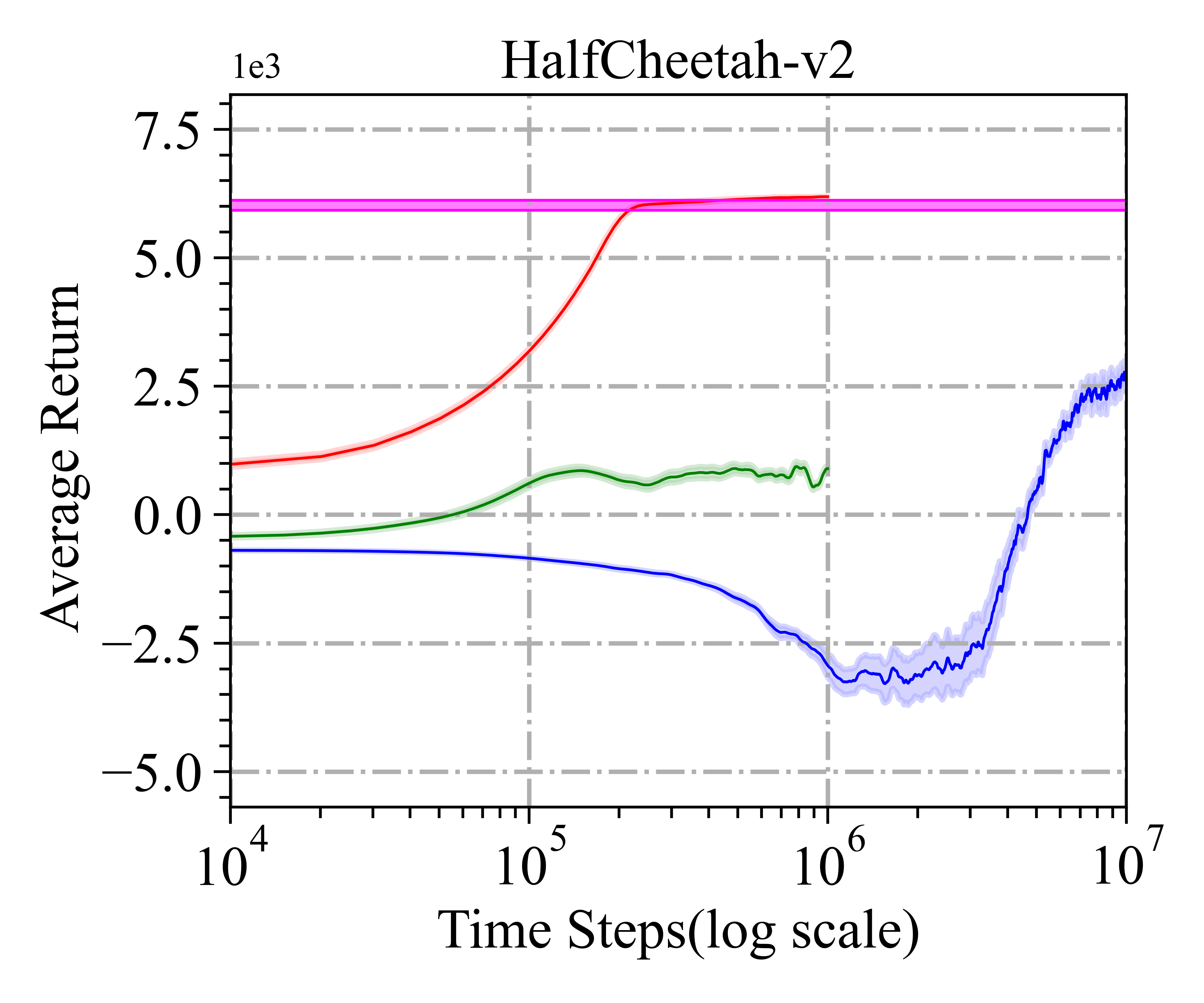}}
	\subfigure{\label{sub.3}
		\includegraphics[width=0.3\linewidth]{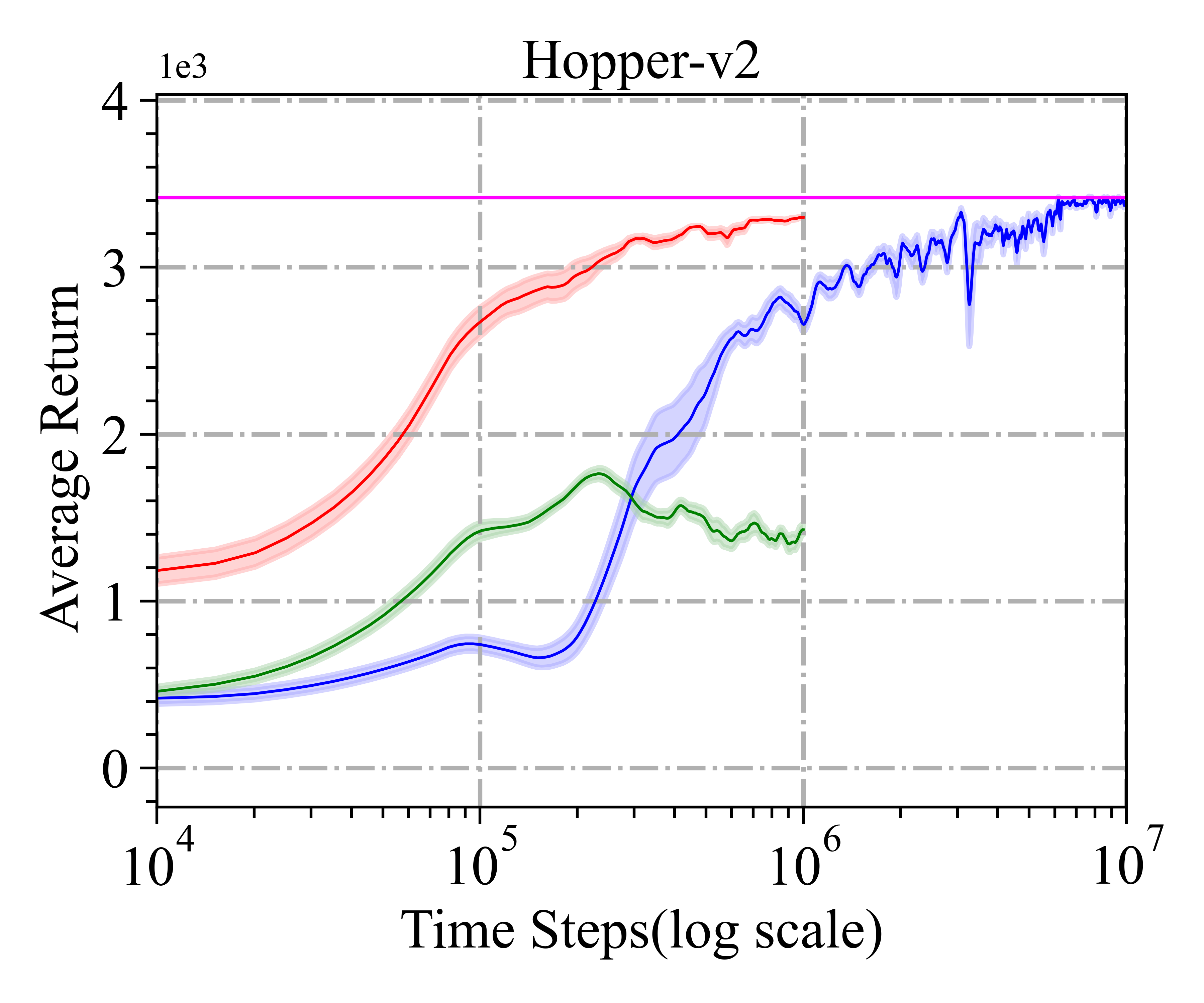}}	
	\subfigure{\label{sub.4}
		\includegraphics[width=0.3\linewidth]{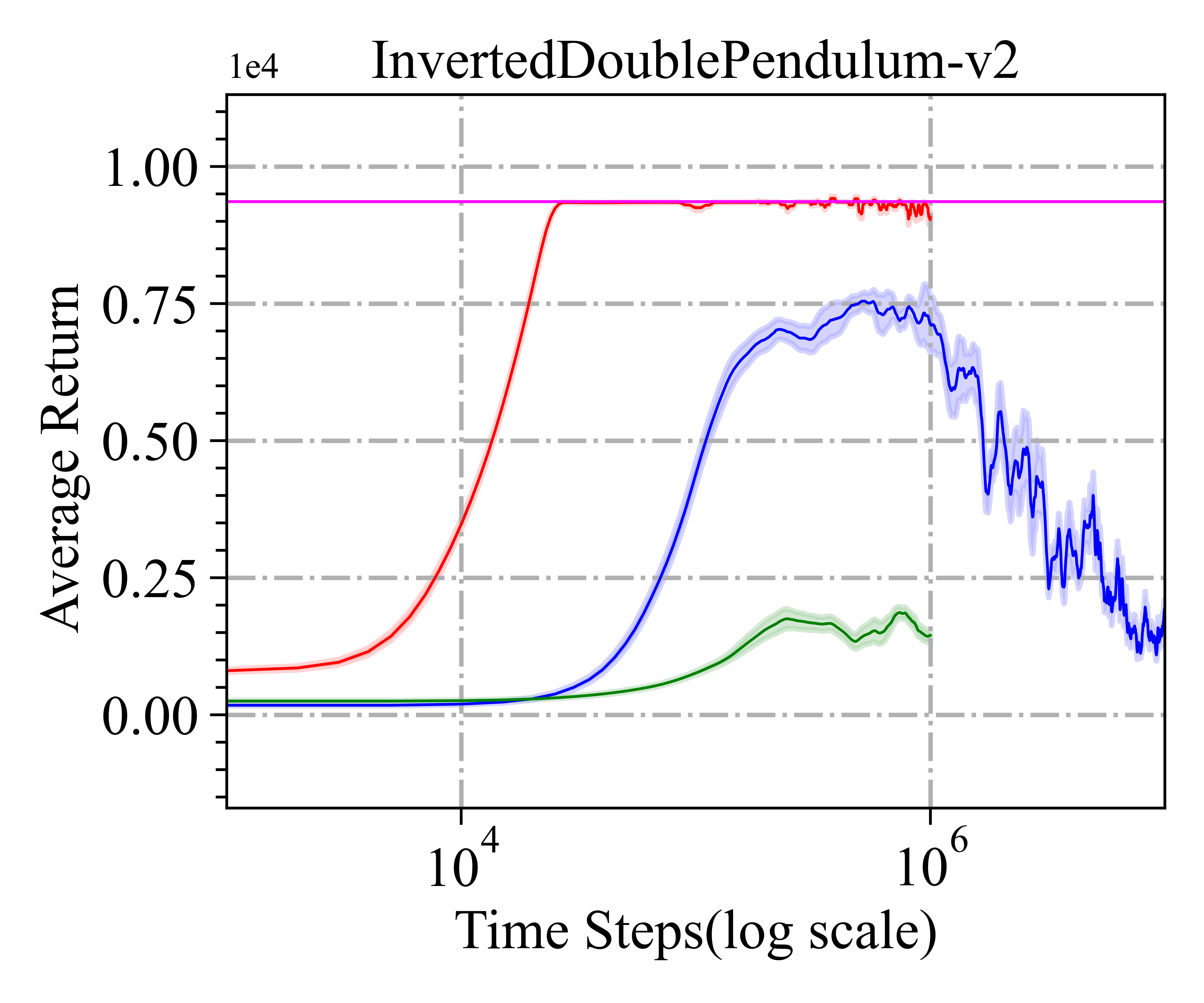}}	
	\subfigure{\label{sub.5}
		\includegraphics[width=0.3\linewidth]{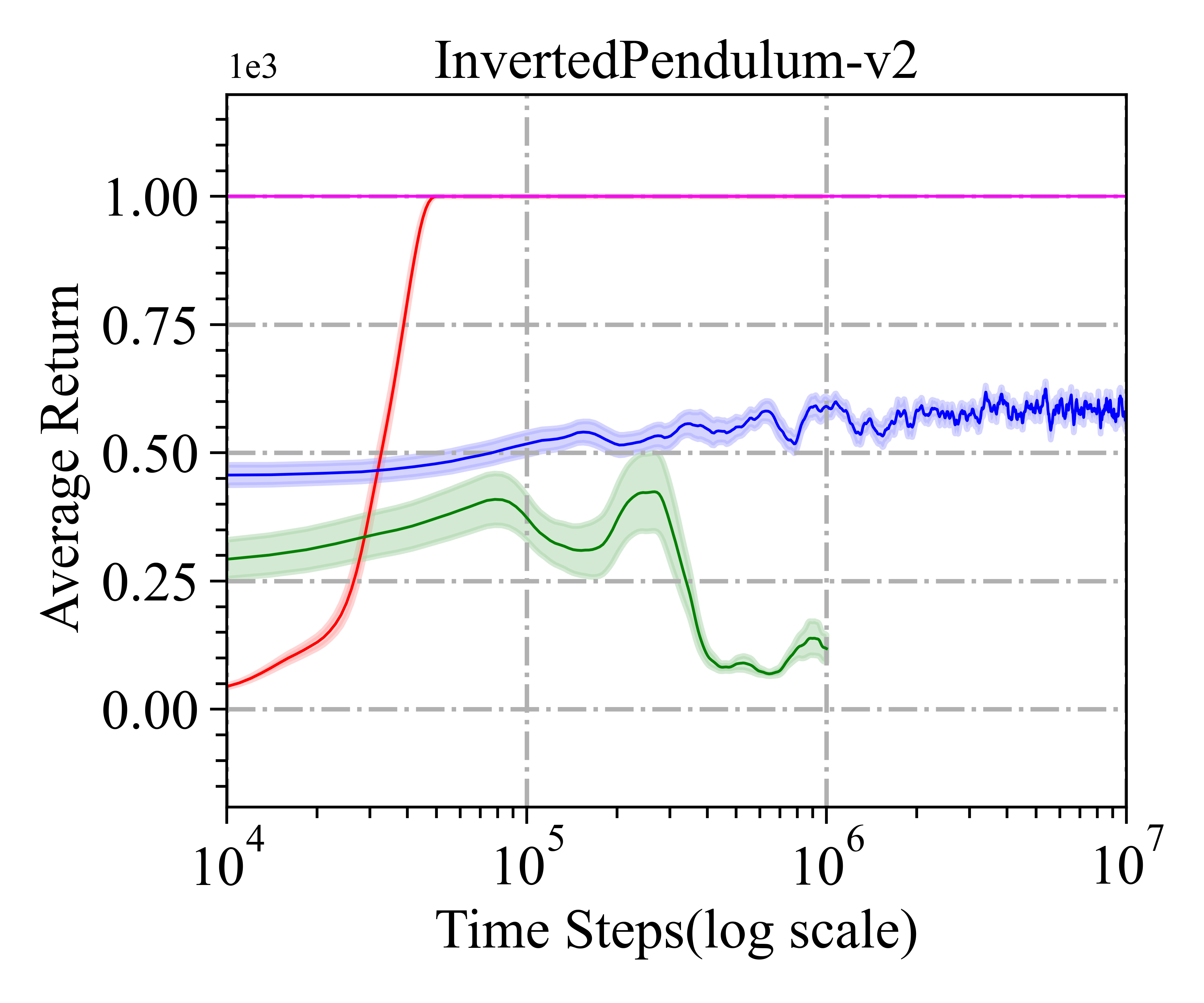}}	
	\subfigure{\label{sub.6}
		\includegraphics[width=0.3\linewidth]{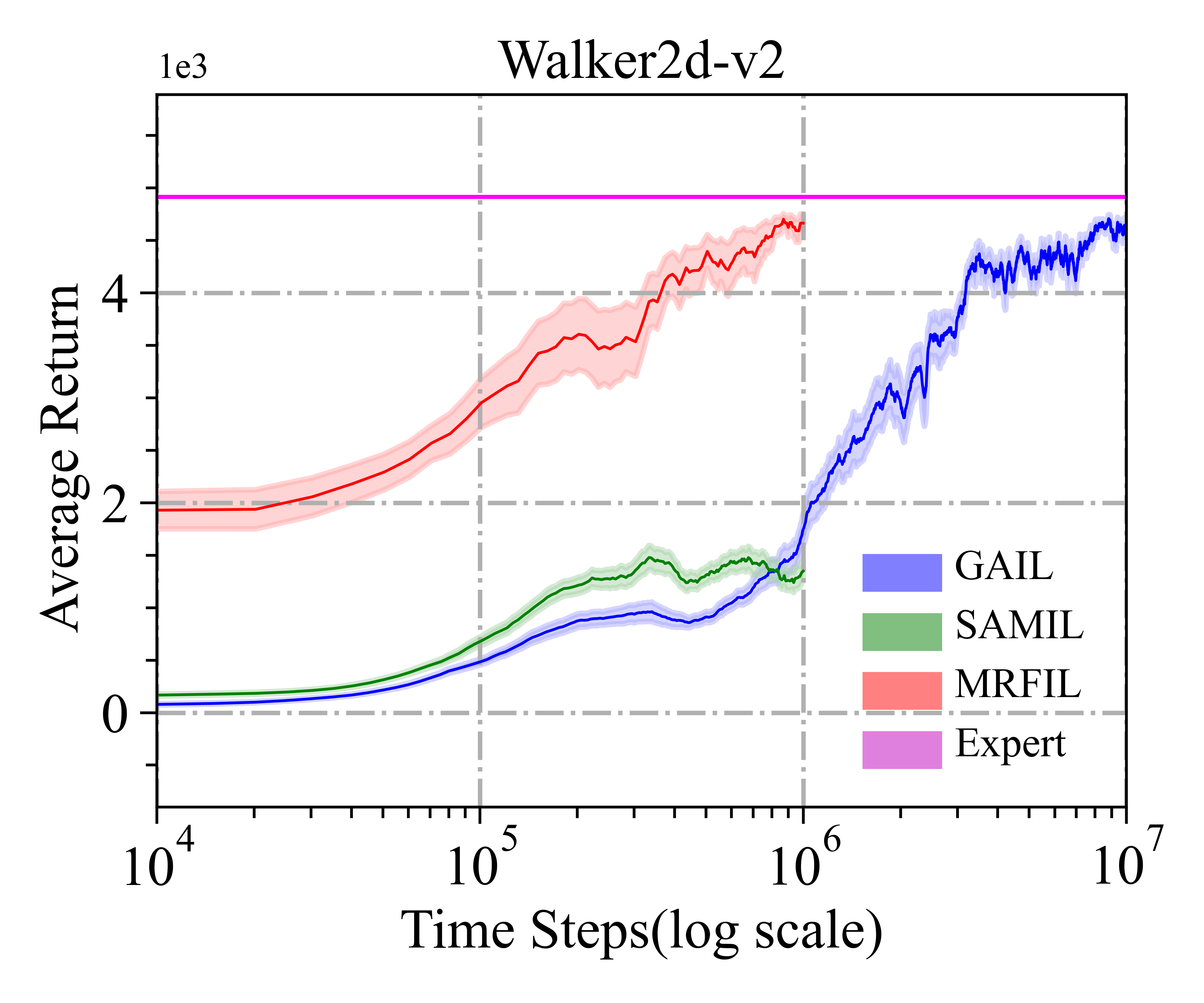}}
	\caption{\textbf{Image-based MuJoCo}. Performance comparison between MRFIL, GAIL and SAMIL in terms of episodic return. The horizontal axis depicts, in logarithmic scale, the number of interactions with environment. Each algorithm is run across 5 random seeds.}
	\label{6Figs}
\end{figure*}

\textbf{Theorem 3}. \textit{The dual problem of MRFIL objective function (Eq (9)) is as follows}:
		\begin{equation} \label{10}
			\begin{split}
				\mathop{\mathrm{minimize}}_{\rho\in D} & \quad -H(\rho) \\
				\mathrm{s.t.} & \quad \rho(s,a) \geq 0 \quad \forall s \in S,\ a \in A \\
				& \quad \pi(s,a)=\pi_{E}(a|s) \quad \forall (s,a)\in D_E	
			\end{split}		
		\end{equation}
\textit{Proof}. See Appendix B, Theorem B.3.\\
 In Theorem 3, the supervised term in Eq(\ref{9})) is converted to $ \pi(s,a)=\pi_{E}(a|s) $, this require the learner becoming as optimal as expert. The constraint in Theorem 3 paly a role equal to the constraint in Theorem 1, thus the learner is guaranteed to take action as same as expert in every state visitation. 

\section{Experiments}
In this section, we comparatively evaluate our proposed method on five continuous control tasks which is built with MuJoCo physics engine, and is wrapped via the OpenAI Gym API. we want to investigate two aspect of MRFIL: the effectiveness of learning from expert demonstrations, and the accuracy of reward function. Specifically, inorder to show its learning performance, we compare MRFIL to two algorithms: GAIL and Sample Efficient Imitation Learning (SAMIL) \cite{DBLP:conf/aistats/BlondeK19}. The goal of this experiment is that investigate not only how well each method can mimic the expert demonstrations, but also how well MRFIL can reduces the interaction numbers.

\subsection{Experimental Settings}
All experiments in this section are conducted in a cluster with two machines with 2 NVIDIA Tesla P40 GPUs each. We implement our algorithms with PyTorch. The agent always starts from the origin point, MRFIL performs on all environments within a 1 million step threshold.\\
We parameterize the actor-critic model using 2-layer ReLU-MLPs and use an ensemble of 5 dynamics models to implement reward function as described in Section 4.3, we parameterize the dynamic model using a 4-layer ReLU-MLPs. For both environments, a Gaussian noise of $ N(0,0.2) $ was added to the states to introduce stochasticity.
\subsection{Empirical Results}
We generate the expert demonstration by SAC that based on Gaussian policy, the expert demonstrations which include 15000 episodes are collected starting from the initial state distribution, and 70 percent of it are used in training process. In the experiment, to simplify the training procedure, the supervised loss term in objective function is computed by calculating the difference betweed BC model and learner, the BC model is trained by expert demonstration. the experiment result demonstrates that this way is feasible.\\
Results of all methods are shown in Figure 1. In these environments we found that MRFIL achieves a higher performance than GAIL and SAMIL in Hopper-v2, Ant-v2 and HalfCheetah-v2. GAIL performs somewhat better than MRFIL on Walker2D-v2. For Ant-v2, Due to the high action dimension, none of GAIL and SAMIL are able to learn optimal policy. The result demonstrates that MRFIL requires significantly less environment interaction on both experimental.\\
Since our approach pre-train the actor-critic network in offline way, and add a supvised loss item. Therefor, these modefications is able to guide the learner to explore nearby expert policy state-action space. Besids, the reward function punishes learner to visit unknown states, thereby providing a safeguard against distribution shift.
\section{Conclusion}
This paper introduced MRFIL, a model-free IL algorithm for continuous control task. The reward function we introduce is acquired by an ensemble of dynamic model from expert demonstrations in offline way. In the sense that the procedure learner interacts with environment is a single RL procedure, this hugely reduce the interaction number. In addition, we propose a model-based offline imitation learning method to pre-train the learner. In the RL procedure, we present a new objective function. Theoretically, we provide a method to measure the return gap between learner and expert when training with model-based IL method. Furthermore, we demonstrate the convergence guarantee for new objective function. Experiment results suggests that MRFIL consistently delivers comparable performance to GAIL and SAMIL, meanwhile MRFIL significantly reduces the environment interactions. In future work, we hope to extend MRFIL to robotic control task.\\

\nocite{DBLP:conf/aistats/BlondeK19}

\appendix
\section{Measure the Policy in Dynamic Model}
In this section we introduce a corollary to measure the return gap between real expert policy and current policy in dynamic model.\\
\textbf{Corollary A 1.} (multi branch and short rollout). Suppose the real dynamic transformation is $ p_r $ , the dynamic model transformation is $ p_m $.

\begin{equation} \label{11}
	\begin{split}
		\left| \eta_e-\eta_m \right| & \leqslant  2\left( \frac{\gamma \left( \epsilon_m+\epsilon_{\pi} \right)}{\left( 1-\gamma \right) ^2}+\frac{\epsilon_{\pi}}{1-\gamma} \right)  \\
		 \text{s.t.} & \quad (s,a) \in \left\{ \left( s,a \right) |r_m\left( s,a \right) =1 \right\} 		
	\end{split}
\end{equation}
\begin{proof}
	Here, $ \eta_{e} $  denotes returns of expert policy $\pi_e$ under real dynamic environment transformation $ p_r $, and $ \eta_{m} $ denotes returns of policy $ \pi_m $ learned by dynamic model $ m_0 $ under dynamics transformation $ p_m $. $ p_e(s,a) $ denotes the state-action pair probability in expert policy $ \pi_e $, $ p_m(s,a) $ denotes the state-action pair probability in current policy $ \pi_m $. $ r(s,a) $ denotes the real reward function, $ r_m(s,a) $ denotes the ensemble reward function. 
	\begin{equation} \label{12}
		\begin{split}
	\mid| \eta _e-\eta _m \mid|=&\left| \sum_{s,a}{\left( p_e\left( s,a \right) -p_m\left( s,a \right) \right)}r_m\left( s,a \right) \right|
	\\
	=&\left| \sum_{s,a}{ \sum_t{\gamma ^t\left( p_{e}^{t}\left( s,a \right) -p_{m}^{t}\left( s,a \right) \right) r_m\left( s,a \right)} } \right|
	\\
	=&\left| \sum_t{\sum_{s,a}{\gamma ^t\left( p_{e}^{t}\left( s,a \right) -p_{m}^{t}\left( s,a \right) \right) r_m\left( s,a \right)}} \right|	
		\end{split}
		\end{equation}
According the reward function (Eq.(\ref{1}) and Eq.(\ref{2})), the reward $ r(s,a) $ must be equal to 1 or 0. Thus we get:
	\begin{equation} \label{13}
		\begin{split}	
	\left| \eta _e-\eta _m \right|=&\left| \sum_t{\sum_{s,a}{\gamma ^t\left( p_{e}^{t}\left( s,a \right) -p_{m}^{t}\left( s,a \right) \right) r_m\left( s,a \right)}} \right|
	\\
	=&\sum_t{\sum_{\left( s,a \right)}{\gamma ^t\left| \left( p_{e}^{t}\left( s,a \right) -p_{m}^{t}\left( s,a \right) \right) \right|}} \\
	\text{s.t.} \quad & \left\{ \left( s,a \right) |r_m\left( s,a \right) =1 \right\} 
		\end{split}
	\end{equation}	
Applying \textbf{Lemma C.2}, using $ \delta=\epsilon_{m}+\epsilon_{\pi}$ (via \textbf{Lemma C.1}), where $\max_{t}\mathbb{E}_{s \sim p_{m}^{t}(s)[D_{KL}(p_{m}(s'|s,a)||p_{e}(s'|s,a))]\leq \epsilon_{m}} $, and $\max_{s} D_{TV}(\pi_{E}(a|s)||\pi_{m}(a|s))\leq \epsilon_{\pi} $. Then:	
	\begin{equation} \label{14}
		\begin{split}	
		D_{TV}\left( p_{m}^{t}\left( s \right) \lVert p_{e}^{t}\left( s \right) \right) \leqslant t\left( \epsilon _m+\epsilon _{\pi} \right) 
		\end{split}
	\end{equation}		
And since we assume $\max_{s} D_{TV}(\pi_{E}(a|s)||\pi_{m}(a|s))\leq \epsilon_{\pi} $, we get:	
		\begin{equation} \label{15}
			\begin{split}	
				\left| \eta _e-\eta _m \right|=
				&\sum_t{\sum_{s,a}{\gamma ^t\left| \left( p_{e}^{t}\left( s,a \right) -p_{m}^{t}\left( s,a \right) \right) \right|}}\\
				\leqslant &2\sum_t{ \gamma ^tt\left( \epsilon _m+\epsilon _{\pi} \right) +\text{e}_{\pi} }\\
				\leqslant& 2\left( \frac{\gamma \left( \epsilon _m+\epsilon _{\pi} \right)}{\left( 1-\gamma \right) ^2}+\frac{\epsilon _{\pi}}{1-\gamma} \right) \\
				\text{s.t.} \quad & \left\{ \left( s,a \right) |r_m\left( s,a \right) =1 \right\} 
			\end{split}
		\end{equation}	
\end{proof}

\section{The Convergence of Objective Function}
In this section we introduce three theorems to demonstrate the convergency of our algorithm.\\
\subsection{Proof of Theorem 1}
\textbf{Theorem 1.} \textit{The dual of inverse reinforcement problem is as follows}:
	\begin{equation} \label{16}
		\begin{split}	
		\underset{\rho \in D}{\min\text{mize}}& \, -H\left( \rho \right) 
		\\
		\,\,   \!\:\!\:\text{s}.\text{t}.&   \!\:\,\,\rho \left( s,a \right) =\rho _E\left( s,a \right) \,\,\forall s\in S,a\in A 
		\end{split}
	\end{equation}
\begin{proof}
	We denote occupancy measure $ \rho_\pi:S\times A\rightarrow\mathbb{R} $ as $\rho_{\pi}\left(s,a\right)=\pi\left(a\middle|s\right)\sum_{t=0}^{\infty}{\gamma^tP\left(s_t=s\middle|\pi\right)} $.
		\begin{equation} \label{17}
			\begin{split}	
				\underset{\rho \in D}{\min\text{mize}}&-H\left( \rho \right) 
				\\
				\,\,   \!\:\!\:\text{s}.\text{t}.&   \!\:\,\,\rho \left( s,a \right) =\rho _E\left( s,a \right) \,\,\forall s\in S,a\in A
			\end{split}
		\end{equation}
Where we denote occupancy measure$ \rho_\pi:S\times A\rightarrow\mathbb{R} $ as $ \rho_\pi\left(s,a\right)=\pi\left(a\middle|s\right)\sum_{t=0}^{\infty}{\gamma^tP\left(s_t=s\middle|\pi\right)} $, we introduce generalized Lagrange function, and asume $ \max_{\alpha}\max_{\rho(s,a)}L_{1}(\rho,c) = \Psi_{1}^{1} $:	
		\begin{equation} \label{18}
			\begin{split}	
		\Psi_{1}^{1} =-H\left( \rho \right)
		+\sum_{s,a}{\alpha _{s,a}\left( \rho _E\left( s,a \right) -\rho \left( s,a \right) \right)}
			\end{split}
		\end{equation}	
Denote the dual variables $ \alpha_{s,a} $ as reward $ r\left(s,a\right) $, and assume $ \max_{r(s,a)}\min_{\rho(s,a)}L_{1}(\rho,c) = \Psi_{1}^{2} $
		\begin{equation} \label{19}
			\begin{split}	
\Psi_{1}^{2} =&-H\left( \rho \right) +\sum_{s,a}{r\left( s,a \right) \left( \rho _E\left( s,a \right)-\rho \left( s,a \right) \right)}\\
=&-H\left( \rho \right) +\sum_{s,a}{r\left( s,a \right) \rho _E\left( s,a \right)} -\sum_{s,a}{r\left( s,a \right) \rho \left( s,a \right)}\\
=&-H\left( \rho \right) + \sum_{s,a}{r\left( s,a \right) \pi _E\left( a|s \right) \sum_t{\gamma ^t}p_{E}^{t}\left( s_t=s|\pi _E \right)}\\
&-\sum_{s,a}{r\left( s,a \right) \pi \left( a|s \right) \sum_t{\gamma ^t}p^t\left( s_t=s|\pi \right)}\\
=&-H\left( \rho \right) +\sum_{s,a}{\pi _E\left( a|s \right) \sum_t{\gamma ^t}p_{E}^{t}\left( s_t=s|\pi _E \right) r\left( s,a \right)}\\
&-\sum_{s,a}{\pi \left( a|s \right) \sum_t{\gamma ^t}p^t\left( s_t=s|\pi \right) r\left( s,a \right)}\\
=&-H\left( \rho \right) -\mathbb{E}_{\pi}\left[ r\left( s,a \right) \right] +\mathbb{E}_{\pi _E}\left[ r\left( s,a \right) \right] 
			\end{split}
		\end{equation}		
Applying \textbf{Lemma C.4}, we get:	
		\begin{equation} \label{20}
			\begin{split}	
		\underset{r\left( s,a \right)}{\max}\underset{\rho \left( s,a \right)}{\min}L_1\left( \rho ,c \right) =-H\left( \rho \right) -\mathbb{E}_{\pi}\left[ r\left( s,a \right) \right] +\mathbb{E}_{\pi _E}\left[ r\left( s,a \right) \right] 
		\\
		=-H\left( \pi \right) -\mathbb{E}_{\pi}\left[ r\left( s,a \right) \right] +\mathbb{E}_{\pi _E}\left[ r\left( s,a \right) \right] 
			\end{split}
		\end{equation}		
\end{proof}

\subsection{Proof of Theorem 2}	
\textbf{Theorem 2.} \textit{The dual problem of policy gradient methods objective function Eq (\ref{7}) is as follows}:	
\begin{equation} \label{21}
	\begin{split}	
		\underset{\rho \in D}{\min\text{mize}}&-H\left( \rho \right) 
		\\
		\,\,   \!\:\!\:\text{s}.\text{t}.&   \!\:\,\,\rho \left( s,a \right) \geqslant 0 ,
		 \!\:\,\, \forall s\in S,a\in A 
	\end{split}
\end{equation}
\begin{proof}
We introduce generalized Lagrange function $ \max_\alpha{\min_{\rho\left(s,a\right)}{L_2\left(\rho,\alpha\right)}}=\Psi_{2}^{1} $:

\begin{equation} \label{22}
	\begin{split}	
		\Psi_{2}^{1} =-H\left( \rho \right) -\sum_{s,a}{\alpha _{s,a}\rho \left( s,a \right)}
	\end{split}
\end{equation}
Denote the dual variables $ \alpha_{s,a} $ as reward$  r\left(s,a\right) $, and $ \alpha_{s,a}\geq0 $, and assume $ \max_{r\left( s,a \right)}\min_{\rho(s,a)}L_{2}(\rho,c) = \Psi_{2}^{2} $
\begin{equation} \label{23}
	\begin{split}	
\Psi_{2}^{2} =&-H\left( \rho \right) -\sum_{s,a}{r\left( s,a \right) \rho \left( s,a \right)}
\\
=&-H\left( \rho \right) -\sum_{s,a}{r\left( s,a \right) \pi \left( a|s \right) \sum_t{\gamma ^t}p^t\left( s_t=s|\pi \right)}
\\
=&-H\left( \rho \right) -\sum_{s,a}{\pi \left( a|s \right) \sum_t{\gamma ^t}p^t\left( s_t=s|\pi \right) r\left( s,a \right)}
\\
=&-H\left( \rho \right) -\mathbb{E}_{\pi}\left[ r\left( s,a \right) \right] 
	\end{split}
\end{equation}
Applying\textbf{ Lemma C.4}, we get:
\begin{equation} \label{24}
	\begin{split}	
\underset{r\left( s,a \right)}{\max}\underset{\rho \left( s,a \right)}{\min}L_2\left( \rho ,c \right) =&-H\left( \rho \right) -\mathbb{E}_{\pi}\left[ r\left( s,a \right) \right] 
\\
=&-H\left( \pi \right) -\mathbb{E}_{\pi}\left[ r\left( s,a \right) \right] 
	\end{split}
\end{equation}
\end{proof}	

\subsection{Proof of Theorem 3}
\textbf{Theorem 3.} The dual problem of MRFIL objective function (Eq (\ref{9})) is as follows:
\begin{equation} \label{25}
	\begin{split}	
\underset{\rho \in D}{\min\text{mize}}&-H\left( \rho \right) 
\\
\,\,   \!\:\!\:\text{s}.\text{t}.&   \!\:\,\,\rho \left( s,a \right) \geqslant 0 ,\!\:\,\,  \forall s\in S,a\in A
\\
\qquad \qquad &\pi \left( a|s \right) =\rho _E\left( a|s \right) ,\,\,\forall \left( s,a \right) \in D_E 
	\end{split}
\end{equation}
\begin{proof}
We introduce generalized Lagrange function $ \max_{\alpha,\beta}{\min_{\rho\left(s,a\right)}{L_3\left(\rho,\alpha,\beta\right)}}=
\Psi_{3}^{1}$:
\begin{equation} \label{26}
	\begin{split}	
		\Psi_{3}^{1} =&-H\left( \rho \right) +\sum_{s,a}{\alpha _{s,a}\left( \rho _E\left( s,a \right) -\rho \left( s,a \right) \right)}\\
		&-\sum_{s,a}{\beta _{s,a}\rho \left( s,a \right)}
	\end{split}
\end{equation}
Denote the dual variables $ \beta_{s,a} $ as reward $ r\left(s,a\right) $, and $ \beta_{s,a}\geq0 $, and assume $ \max_{r\left( s,a \right)}\min_{\pi(s,a)}L_{3}(\rho,r,\alpha) = \Psi_{3}^{2} $
\begin{equation} \label{27}
	\begin{split}	
\Psi_{3}^{2} =&-H\left( \rho \right) -\sum_{s,a}{r\left( s,a \right) \rho \left( s,a \right)}\\
&+\sum_{\left( s,a \right) \in D_E}{\alpha _{s,a}\left( \pi _E\left( a|s \right) -\pi \left( a|s \right) \right)}
\\
=&-H\left( \rho \right) -\sum_{s,a}{r\left( s,a \right) \pi \left( a|s \right) \sum_t{\gamma ^t}p^t\left( s_t=s|\pi \right)}\\
&+\sum_{\left( s,a \right) \in D_E}{\alpha _{s,a}\left( \pi _E\left( a|s \right) -\pi \left( a|s \right) \right)}
\\
=&-H\left( \rho \right) -\sum_{s,a}{\pi \left( a|s \right) \sum_t{\gamma ^t}p^t\left( s_t=s|\pi \right) r\left( s,a \right)}\\
&+\sum_{\left( s,a \right) \in D_E}{\alpha _{s,a}\left( \pi _E\left( a|s \right) -\pi \left( a|s \right) \right)}
\\
=&-H\left( \rho \right) -\mathbb{E}_{\pi}\left[ r\left( s,a \right) \right]\\
 &+\sum_{\left( s,a \right) \in D_E}{\alpha _{s,a}\left( \pi _E\left( a|s \right) -\pi \left( a|s \right) \right)}
	\end{split}
\end{equation}
When policy distribution $ \pi\left(s,a\right) $ and $ \pi_E\left(s,a\right) $ is very close, the dual parameters $ \alpha_{s,a} $ have minimal impact on $ L_3\left(\rho,r,\alpha\right) $, which can be ignored. We random sample state-action pair in expert demonstration, and take $ \alpha_{s,a} $ as sample probability, and assume $ \max_{r\left( s,a \right)}\min_{\pi(s,a)}L_{3}(\rho,r,\alpha) = \Psi_3 $. Therefore:
\begin{equation} \label{28}
	\begin{split}	
\underset{r\left( s,a \right)}{\max}\underset{\rho \left( s,a \right)}{\min}L_3\left( \rho ,c \right) =&-H\left( \rho \right) -\mathbb{E}_{\pi}\left[ r\left( s,a \right) \right]\\
 &+\sum_{\left( s,a \right) \in D_E}{\alpha _{s,a}\left( \pi _E\left( s,a \right) -\pi \left( s,a \right) \right)}
\\
\thickapprox& -H\left( \rho \right) -\mathbb{E}_{\pi}\left[ r\left( s,a \right) \right]\\ &+\mathbb{E}_{D_E}\left[ \pi _E\left( s,a \right) -\pi \left( s,a \right) \right] 
	\end{split}
\end{equation}
\end{proof}

\subsection{Useful Lemmas}
In this section we introduce two useful lemmas used in our corollary. which is provided in \cite{DBLP:conf/nips/JannerFZL19}.\\
\textbf{Lemma C.1} (TVD of Joint Distributions.) \textit{Suppose we have two distributions} $ p_1\left(x,y\right)=p_1\left(x\right)p_1\left(y\middle| x\right) $ and $ p_2\left(x,y\right)=p_2\left(x\right)p_2\left(y\middle| x\right) $. \textit{We assume the total variation distance} $ D_{TV}\left( p_1\left( x,y \right) \parallel p_2\left( x,y \right)\right)$ as $ \varPhi $, \textit{we can bound the total variation distance of the joint as}:
\begin{equation} \label{29}
	\begin{split}	
		\varPhi \leqslant& D_{TV}\left( p_1\left( x \right) 
		\parallel p_2\left( x \right) \right)\\ 
		+&\mathbb{E}_{x~p_1}\left[ D_{TV}\left( p_1\left( y|x \right) \parallel p_2\left( y|x \right)\right) \right] 
	\end{split}
\end{equation}

\begin{proof}
\begin{equation} \label{30}
	\begin{split}	
  		\varPhi =&\frac{1}{2}\sum_{x,y}{\left| p_1\left( x,y \right) -p_2\left( x,y \right) \right|}
  		\\
  		=&\frac{1}{2}\sum_{x,y}{\left| p_1\left( x \right) p_1\left( y|x \right) -p_2\left( x \right) p_2\left( y|x \right) \right|}
  		\\
  		=&\frac{1}{2}\sum_{x,y} \bigl( \mid p_1\left( x \right) p_1\left( y|x \right) -p_1\left( x \right) p_2\left( y|x \right) \\
  			&+\left( p_1\left( x \right) -p_2\left( x \right) \right) p_2\left( y|x \right) \mid \bigr)
  		\\
  		\leqslant& \frac{1}{2}\sum_{x,y} \bigl( p_1\left( x \right) \left| p_1\left( y|x \right) -p_2\left( y|x \right) \right|\\
  		&+\left| p_1\left( x \right) -p_2\left( x \right) \right|p_2\left( y|x \right) \bigr)
  		\\
  		=&\frac{1}{2}\sum_{x,y}{p_1\left( x \right) \left| p_1\left( y|x \right) -p_2\left( y|x \right) \right|}\\
  		 &+\frac{1}{2}\sum_x{\left| p_1\left( x \right) -p_2\left( x \right) \right|}
  		\\
  		=&\mathbb{E}_{x~p_1}\left[ D_{TV}\left( p_1\left( y|x \right) \lVert p_2\left( y|x \right) \right) \right]\\ 
  		 &+D_{TV}\left( p_1\left( x \right) \lVert p_2\left( x \right) \right) 
  		\\
  		\leqslant& \underset{x}{\max}D_{TV}\left( p_1\left( y|x \right) \lVert p_2\left( y|x \right) \right)\\ 
  		 &+D_{TV}\left( p_1\left( x \right) \lVert p_2\left( x \right)  \right) 
	\end{split}
\end{equation}
\end{proof}
\setlength{\parindent}{0em}\textbf{Lemma C.2} (Markov chain TVD bound, time-varying). \textit{Suppose the expert KL-divergence between two transition distribution is bounded as  and the initial state distribution are the same} $ p_1^{t=0}\left(s\right)=p_2^{t=0}\left(s\right) $. \textit{Then the distance in the state marginal is bounded as}:
\begin{equation} \label{31}
	\begin{split}	
		D_{TV}\left( p_{1}^{t}\left( s \right) \parallel p_{2}^{t}\left( s \right) \right) \leqslant t\delta  
	\end{split}
\end{equation}
\begin{proof}
	We begin by bounding the TVD in state-visitation at time $ t $, which is denoted as$  \epsilon_t=(p_1^t(s)||p_2^t\left(s\right))$, and assume $ \left| p_{1}^{t}\left( s \right) -p_{2}^{t}\left( s \right) \right| $ as $\varGamma$:
	\begin{equation} \label{32}
		\begin{split}
		\varGamma=&\left| \sum_{s'}{p_1\left( s_t=s|s' \right) p_{1}^{t-1}\left( s' \right) -p_2\left( s_t=s|s' \right) p_{2}^{t-1}\left( s' \right)} \right|
		\\
		\leqslant& \sum_{s'}{\left| p_1\left( s_t=s|s' \right) p_{1}^{t-1}\left( s' \right) -p_2\left( s_t=s|s' \right) p_{2}^{t-1}\left( s' \right) \right|}
		\\
		=&\sum_{s'} \Bigl[ \mid p_1\left( s|s' \right) p_{1}^{t-1}\left( s' \right) -p_2\left( s|s' \right) p_{1}^{t-1}\left( s' \right)\\
		 &+p_2\left( s|s' \right) p_{1}^{t-1}\left( s' \right) -p_2\left( s|s' \right) p_{2}^{t-1}\left( s' \right) \mid \Bigr]  \\
		\leqslant& \sum_{s'} \Bigl[ {p_{1}^{t-1}\left( s' \right) \left| p_1\left( s|s' \right) -p_2\left( s|s' \right) \right|}\\
		 &+p_2\left( s|s' \right) \left| p_{1}^{t-1}\left( s' \right) -p_{2}^{t-1}\left( s' \right) \right| \Bigr]
		\\
		=&\mathbb{E}_{s'~p_{1}^{t-1}}\Bigl[ \left| p_1\left( s|s' \right) -p_2\left( s|s' \right) \right| \Bigr]\\ 
		 &+\sum_{s'}{p\left( s|s' \right) \left| p_{1}^{t-1}\left( s' \right) -p_{2}^{t-1}\left( s' \right) \right|}
		\end{split}
	\end{equation}
\begin{equation} \label{33}
	\begin{split}	
		\epsilon _t=&D_{TV}\left( p_{1}^{t}\left( s \right) \lVert p_{2}^{t}\left( s \right) \right) 
		\\
		=&\frac{1}{2}\sum_s{\left| p_{1}^{t}\left( s \right) -p_{2}^{t}\left( s \right) \right|}
		\\
		=&\frac{1}{2}\sum_s{\left| p_{1}^{t}\left( s \right) -p_{2}^{t}\left( s \right) \right|}\Bigl( \mathbb{E}_{s'~p_{1}^{t}}\bigl[ \left| p_1\left( s|s' \right) -p_2\left( s|s' \right) \right| \bigr]\\
		 &+D_{TV}\left( p_{1}^{t-1}\left( s' \right) \parallel p_{2}^{t-1}\left( s' \right) \right) \Bigr) 
		\\
		=&\delta _t + \epsilon _{t-1}
		\\
		=&\epsilon _0+\sum_{i=0}^t{\delta _i}\\
		=&\sum_{i=0}^t{\delta _i}\\
		=&t\delta 
	\end{split}
\end{equation}
Where we have defined $ \delta_t=\frac{1}{2}\sum_{s'\sim p_1^{t-1}}[\sum_{s}\mid p_{1}(s \mid s')-p_{2}(s\mid s')\mid] $, which we assume is upper bounded by $ \delta $. Assuming we are not modeling the initial state distribution, we can set $ \epsilon_0=0 $.
\end{proof}
\textbf{Lemma C.3} (Theorem 2 of \cite{DBLP:conf/icml/SyedBS08}). \textit{if} $ \rho\in D $, \textit{then} $ \rho $ \textit{is the occupancy measure for} $ \pi_\rho\left(a\middle| s\right)\triangleq\frac{\rho\left(s,a\right)}{\sum_{a^\prime}\rho\left(s,a^\prime\right)} $, \textit{and} $ \pi_\rho $ \textit{is the only policy whose occupancy measure is} $ \rho $.\\
\textbf{Lemma C.4} (Lemma 3.2 of \cite{DBLP:conf/nips/HoE16}) \textit{Let} $ \bar{H}\left(\rho\right)=-\sum_{s,a}{\rho\left(s,a\right)\log{(\rho(s,a)/\sum_{a^\prime}{\rho\left(s,a^\prime\right)).\ }}} $ \textit{Then}, $ \bar{H}\ $ \textit{is strictly concave, and for all} $ \pi\in\prod $ \textit{and} $ \rho\in D $, \textit{we have} $ H\left(\pi\right)=\bar{H}\left(\rho_\pi\right) $ and $ \bar{H}\left(\rho\right)=H\left(\pi_\rho\right) $.

\label{stylefiles}

\bibliographystyle{kr}
\bibliography{reference}

\end{document}